\documentclass[conference]{IEEEtran}
\IEEEoverridecommandlockouts
\usepackage{cite}
\usepackage{amsmath,amssymb,amsfonts}
\usepackage{algorithmic}
\usepackage{graphicx}
\usepackage{textcomp}
\usepackage{xcolor}
\usepackage{hyperref}
\def\BibTeX{{\rm B\kern-.05em{\sc i\kern-.025em b}\kern-.08em
    T\kern-.1667em\lower.7ex\hbox{E}\kern-.125emX}}
\begin{document}

\title{A Transfer Learning Approach for Dialogue Act Classification of GitHub Issue Comments\\
\thanks{This research was supported by DARPA program HR001117S0018.}
}

\author{\IEEEauthorblockN{Ayesha Enayet}
\IEEEauthorblockA{\textit{Department of Computer Science} \\
\textit{University of Central Florida}\\
Orlando, FL US \\
ayeshaenayet@knights.ucf.edu}
\and
\IEEEauthorblockN{Gita Sukthankar}
\IEEEauthorblockA{\textit{Department of Computer Science} \\
\textit{University of Central Florida}\\
Orlando, FL US \\
gitars@eecs.ucf.edu}

}

\maketitle

\begin{abstract}
Social coding platforms, such as GitHub, serve as  laboratories for studying collaborative problem solving in open source software development; a key feature is their ability to support issue reporting which is used by teams to discuss tasks and ideas.  Analyzing the dialogue between team members, as expressed in issue comments, can yield important insights about the performance of virtual teams.  This paper presents a transfer learning approach for performing dialogue act classification on issue comments.  Since no large labeled corpus of GitHub issue comments exists, employing transfer learning enables us to leverage standard dialogue act datasets in combination with our own GitHub comment dataset. We compare the performance of several word and sentence level encoding models including Global Vectors for Word Representations (GloVe), Universal Sentence Encoder (USE), and Bidirectional Encoder Representations from Transformers (BERT). Being able to map the issue comments to dialogue acts is a useful stepping stone towards understanding cognitive team processes.
\end{abstract}

\begin{IEEEkeywords}
social coding platforms, dialogue act classification, transfer learning, embeddings for natural language processing
\end{IEEEkeywords}

\section{Introduction}
The emergence of online collaboration platforms has dramatically changed the dynamics of human teamwork, creating a veritable army of virtual teams, composed of workers in different physical locations. Software engineering requires a tremendous amount of collaborative problem solving, making it an excellent domain for team cognition researchers who seek to understand the manifestation of cognition applied to team tasks.  Mining data from social coding platforms such as GitHub can yield insights about the thought processes of virtual teams.  Previous work on issue comments~\cite{guzman2014,murgia2018,ortu2018} has focused on emotional aspects of team communication, such as sentiment and politeness.  Our aim is to map issue comments to states in team cognition such as information gathering, knowledge building and problem solving.  To do this we employ dialogue act (DA) classification, in order to identify the intent of the speaker.

Dialogue act classification has a broad range of natural language processing applications, including machine translation, dialogue systems and speech recognition.  Semantic-based classification of human utterances is a challenging task, and the lack of a large annotated corpus that represents class variations makes the job even harder. Compared to the examples of human utterances available in standard datasets like the Switchboard (SwBD) corpus and the CSI Meeting Recorder Dialogue Act (MRDA) corpus, GitHub utterances are more complex. 

The primary purpose of our study is the DA classification of GitHub issue comments by harnessing the strength of transfer learning, using word and sentence level embedding models fine-tuned on our dataset.  For word-level transfer learning, we have used GLoVe vectors~\cite{pennington2014glove}, and Universal Sentence Encoders~\cite{cer2018universal} and BERT~\cite{devlin2018bert} models were used for sentence-level transfer. This paper presents a comparison of the performance of various  architectures on GitHub dialogues in a limited resource scenario.  A second contribution is our publicly available dataset of annotated issue comments. The dataset is available at \url{https://drive.google.com/drive/folders/1kLZvzfE80VeEYA1tqua_aj6nSiT57f83?usp=sharing}.
In the field of computational collective intelligence,  where people collaborate and work in teams to achieve goals, dialogue act classification can play a vital role in understanding human teamwork. 

\section{Background}
Unlike general purpose communication platforms such as Twitter and Facebook, GitHub is specialized to support virtual teams of software developers whose primary communication goal is to discuss new features and monitor software bugs. It facilitates distributed, asynchronous collaborations in open source software (OSS) development.   Code development, issue reporting, and social interactions are tracked by the 20+ event types. Our assumption is that each software repository is maintained by a team and that the events associated with the repository form a partial history of the team activities and social interactions. 

GitHub has an open API to collect metadata about users, repositories, and the activities of users on repositories. Developers make changes to the code repository by pushing their content, while GitHub tracks the version control process. Any GitHub user can contribute to a repository by sending a pull request. Repository maintainers review pull requests, discuss possible modifications in the comments, and decide whether to accept or reject the requests. GitHub also supports passive social media style interactions such as following repositories or developers.  Within GitHub’s issue handling infrastructure, users can report a bug or provide a feature request by opening an issue.  Issue closure rates thus reflect the speed with which teams resolve problems and can be used as a measure of team performance. 
\section{Related Work}
Issue resolution has been viewed by many researchers as a rich source of information about the emotional health of the team and how it affects the software development process~\cite{islam2016}.  Kikas et al. demonstrated a model for predicting issue lifetime that included a single feature aggregating textual comment information~\cite{kikas2016}. 
Several studies have employed sentiment analysis~\cite{guzman2014,murgia2018,ortu2018} and topic modeling~\cite{wang2019} to study GitHub issue comments. Ortu et al. conducted a large study on communication patterns in which they measured politeness and emotional affect in issue comments; their aim was to understand how contribution levels modulate communication patterns~\cite{ortu2018}.  Murgia et al. demonstrated a  machine learning classifier for identifing love, joy or sadness in issue comments~\cite{murgia2018}. An empirical study of issue comments conducted by Guzman et al.~\cite{guzman2014} showed that the sentiment expressed in issue comments varies based on day of week, geographic dispersion of the team, and the programming language. Yang et al. addressed the more practical question of the relationship of issue comment sentiment and bug fixing speed~\cite{yang2017}. 

In contrast, our aim is to study the team cognition aspects of collaborative problem solving using dialogue act classification.  Unlike topic modeling or sentiment analysis, dialogue act classification has not been extensively applied to GitHub data.  However, Saha et al.~\cite{saha2019tweet} proposed a deep learning approach for the dialogue act  classification of Twitter data. A convolutional neural network was used to create the classifier, along with hand-crafted rules.  Seven classes were included: statement, expression, suggestion, request, question, threat, and other.  In contrast, our work is done using a transfer learning approach and a significantly larger set of classes.

Prior to deep learning, statistical approaches such as hidden Markov models, have been used for dialogue act classification. The HMM represents discourse structure, with dialogue acts as states.  Stolcke et al. demonstrated such a model that combined
prosodic, lexical, and collocational cues~\cite{stolcke2000dialogue}.  Chen et al. proposed the CRF-Attentive Structured Network (CRF-ASN) framework to exploit the CRF-attentive structure dependencies along with end-to-end training~\cite{chen2018dialogue}. 

This paper presents a transfer learning approach for dialogue act classification that is used to compensate for our small dataset.   To do this, we learn an embedding from a larger dataset.
The next section surveys the state of the art in embedding models for natural language processing.
\subsection{Embeddings}
Embeddings are a mechanism for mapping a high-dimensional space to a low-dimensional one while only retaining the most effective structural representations. They can be used as part of the transfer learning process to mitigate the low availability of labeled language resources on various NLP tasks.  This paper presents transfer learning results using the following state of the art embedding methods: Global Vectors for Word Representation (GloVe)~\cite{pennington2014glove}, Universal Sentence Encoding (USE)~\cite{cer2018universal}, and Bidirectional Encoding (BERT)~\cite{devlin2018bert}.  We compare these embedding models with the probabilistic technique proposed by Duran et al.~\cite{duran2018probabilistic} on our GitHub issue comments dataset. 
\subsubsection{Global Vectors for Word Representation (GloVe)}
Pennington et al.~\cite{pennington2014glove} proposed the GloVe model in 2014. It creates a word-level embedding that leverages both the local context window and global matrix factorization methods. GloVe employs a log-bilinear prediction-based technique that utilizes word-word co-occurrence statistics to identify a meaningful structure and generate word-level embeddings. We are using the GloVe model to illustrate the results of DA classification of GitHub data using word-level embedding.
\subsubsection{Universal Sentence Encoders (USE)}
In 2018, Google Research released a Universal Sentence Encoder (USE) model for sentence-level transfer learning that  achieves consistent performance across multiple NLP tasks~\cite{cer2018universal}. There are two different variants of the model: 1) a transformer architecture, which gives high accuracy at the cost of high resource consumption and 2) a deep averaging network that requires few resources and makes small compromises for efficiency. The former uses attention-based, context-aware encoding sub-graphs for the transfer architecture. The model outputs a 512-dimensional vector. The deep averaging network works by averaging words and bigram embeddings to use as an input to a deep neural network. The models are trained on web news, Wikipedia, web question-answer pages, discussion forums, and the Stanford Natural Language Inference (SNLI) corpus. 

\subsubsection{Bidirectional Encoder Representations from Transformers (BERT)}
Also created at Google, BERT is the first model that was trained on both left and right contexts~\cite{devlin2018bert}. To achieve pre-trained deep bidirectional representation, it uses the masked model, which follows the cloze deletion task. This model is trained on Books Corpus and English Wikipedia corpus. The code for BERT is available at \url{https://github.com/ google-research/bert}. There are two available flavors of BERT: 1)$BERT_{BASE}$, and 2) $BERT_{LARGE}$. $BERT_{BASE}$ has 12 transformer blocks, 768 hidden layers, 12 self-attention heads, and 110 million parameters. On the other hand, $BERT_{LARGE}$ uses a fairly large network, with 24 transformer blocks, 1024 hidden layers, 16 self-attention heads, and 340 million parameters.

\section{Method}
Treating our dialogue act classification as a transfer learning problem enables us to leverage embeddings learned on a dataset that is over 200 times larger than our test dataset. Five dialogue act classification pipelines were created to evaluate the performance of four different word and sentence-level embedding models.
\begin{figure*}
  \includegraphics[width=\textwidth,height=18cm]{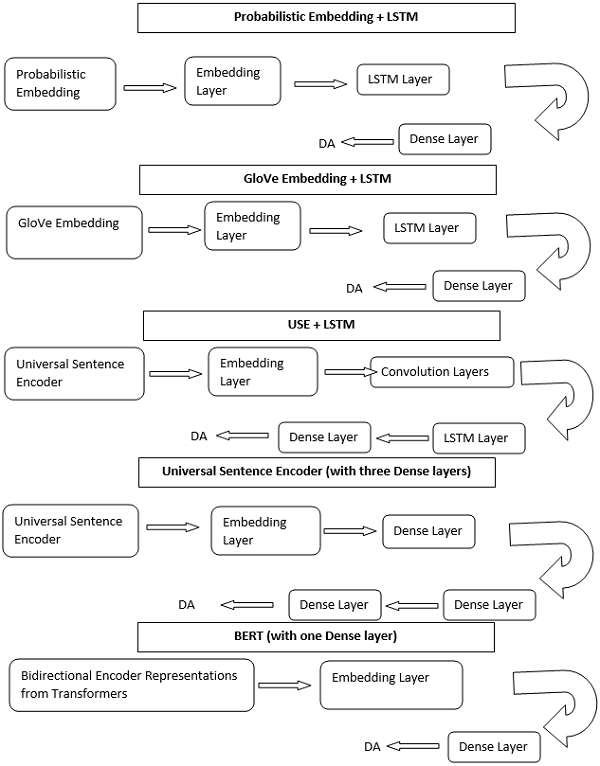}
  \caption{Architecture Diagrams for (i) Probabilistic Representation with RNN (ii) GloVe+LSTM (iii) Universal Sentence Encoder+LSTM (USE+LSTM) (iv) Universal Sentence Encoder (USE) (v) Bidirectional Encoder Representations from Transformers (BERT)}
  \label{Arch}
\end{figure*}
\subsection{Datasets}
For our study, we collected a dataset of issue comments from GitHub and hand annotated them using a standard dialogue act tagset, DAMSL (Discourse Annotation and Markup System of Labeling), to facilitate the transfer process.  We have made the dataset available at \url{https://drive.google.com/drive/folders/1kLZvzfE80VeEYA1tqua_aj6nSiT57f83?usp=sharing}. The tagset is available at \url{https://web.stanford.edu/~jurafsky/ws97/manual.august1.html}.   Our test set consists of 859 instances from more than 50 GitHub issues.  

The models were trained using the  Switchboard Dialogue Act Corpus (SwDA) dataset. SwDA is one of the most popular public datasets for DA classification. It consists of 1155 human-to-human telephone speech conversations. The dataset is tagged using 42 tags from the DAMSL tagset.  Table \ref{tab:desdataset} shows the statistics of both test and train datasets.
Table~\ref{tab:SwDAUtterences} shows examples from the SwDA training dataset. Table~\ref{tab:GitHubUtterences} shows examples from our GitHub issue comment dataset. GitHub issue comments are a complex combination of computer commands, special symbols, and standard English. From these examples, it is clear that this is a challenging transfer learning problem. 
\begin{table}
\centering
\caption{Dataset Statistics}\label{tab:desdataset}
\begin{tabular}{|l|l|l|l|}
\hline
Dataset&Categories& \#Utterances& \#Tokens\\[3ex]
\hline
SwDA (Train) & 42& 200,052&19K\\
GitHub (Test) & 42 & 859& 10,131\\
\hline
\end{tabular}
\end{table}
\begin{table}
\centering
\caption{SwDA Dataset Sample}\label{tab:SwDAUtterences}
\begin{tabular}{|p{1cm}|p{3cm}|p{0.5cm}|p{2cm}|}
\hline
Speaker&Utterance&DA&Description\\[3ex]
\hline
A&I don't, I don't have any kids. & sd& Statement-non-Opinion\\
     A&I, uh, my sister has a, she just had a baby, &sd& Statement-non-Opinion \\
     A& he's about five months old&sd& Statement-non-Opinion \\
     A&and she was worrying about going back to work and what she was going to do with him and -- &sd& Statement-non-Opinion \\
     A&Uh-huh. & b& Acknowledge\\
     A&do you have kids? &qy&Yes-No-Question \\
     B&I have three. &na&Affirmative non-yes Answer \\
     A&Oh, really? &bh&Backchannel in question form \\
\hline
\end{tabular}
\end{table}
\begin{table}
\centering
\caption{GitHub Dataset Sample}\label{tab:GitHubUtterences}
\begin{tabular}{|p{1cm}|p{3cm}|p{0.25cm}|p{2cm}|}
\hline
Speaker&Utterance&DA& Description\\[3ex]
\hline
     A&What steps will reproduce the problem?&qw&Wh-Question\\
     B&Give the *exact* arguments passedto include-what-you-use, and attach the input source file that gives theproblem (minimal test-cases are much appreciated!)&ad&Action-directive \\
     B&Run IWYU against the following file:&ad&Action-directive \\
     B&iwyu  clear& ad&Action-directive\\
     B&iwyu  cat ./cstdarg.cpp&ad&Action-directive\\
     C&What is the expected output?&qw&Wh-Question \\
     C&What do you see instead? &qw&Wh-Question \\
\hline
\end{tabular}
\end{table}
\subsection{Probabilistic Representation with Recurrent Neural Networks}
Duran et al. proposed a probabilistic technique to represent utterances while using the LSTM sentence model for dialogue act classification~\cite{duran2018probabilistic}. The probabilistic distribution of each word in the corpus over DA categories provides the  representation of the utterances. The model does not incorporate contextual features at the discourse level. The set of keywords consisting of all the words that occur above a threshold frequency is used to define a $n\times m$ matrix  X, where $m$ is the number of categories, and $n$ is the number of keywords. Each entry $x_{ij}$ of the matrix represents the probability of the tag $j$ given the word $i$. Training was accomplished using code downloaded from \url{https://github.com/NathanDuran/Probabilistic-RNN-DA-Classifier}.

\subsection{GloVe + LSTM}
We use glove.6b.100d.txt downloaded from \url{https://nlp.stanford.edu/projects/glove/} to train our model on the SwDA dataset. The model consists of input, embedding, LSTM, and one dense layer with 42 output labels.  The network was trained using an Adam optimizer. 

\subsection{Universal Sentence Encoder (USE)}
The Universal Sentence Encoder  was applied to the GitHub issue comments, after fine-tuning on the SwDA dataset. The code to load the USE model is available at \url{https://tfhub.dev/google/universal-sentence-encoder/1}. We chose the USE Transformer-based Architecture model with three dense layers and a softmax activation function.  The network was trained using an Adam optimizer. 
\subsection{USE+LSTM}
We also combine the Universal Sentence Encoder with an LSTM. This model consists of Input, Embedding, Convolution, LSTM, and one Dense output layer.

\subsection{Bidirectional Encoder Representations from Transformers (BERT)}
This architecture was implemented using the $bert\_en\_uncased\_L-12\_H-768\_A-12$ model from TensorFlow Hub. The model has 12 hidden layers (i.e., Transformer blocks), with hidden size of 768, and 12 attention heads.  As proposed by~\cite{devlin2018bert}, we append a single dense layer to BERT.  
\section{Evaluation}
Table \ref{tab:resultCom} shows the performance of all five architectures.  Universal Sentence Encoder had the best performance on the GitHub issue comments, with a test accuracy of 50.71\% which is 6\% more than the accuracy achieved using the probabilistic representation of sentence. The other three models showed significantly lower performance than USE, lagging by almost 10\%. The probabilistic representation of sentence approach exhibited the highest validation accuracy of 76.9\% which is significantly higher than USE which had a validation accuracy of 69.5\%. The well-known BERT model had a validation accuracy of 71.5\%, but had a low test accuracy.
\begin{figure*}[hbt!]
  \includegraphics[width=18cm,height=16cm]{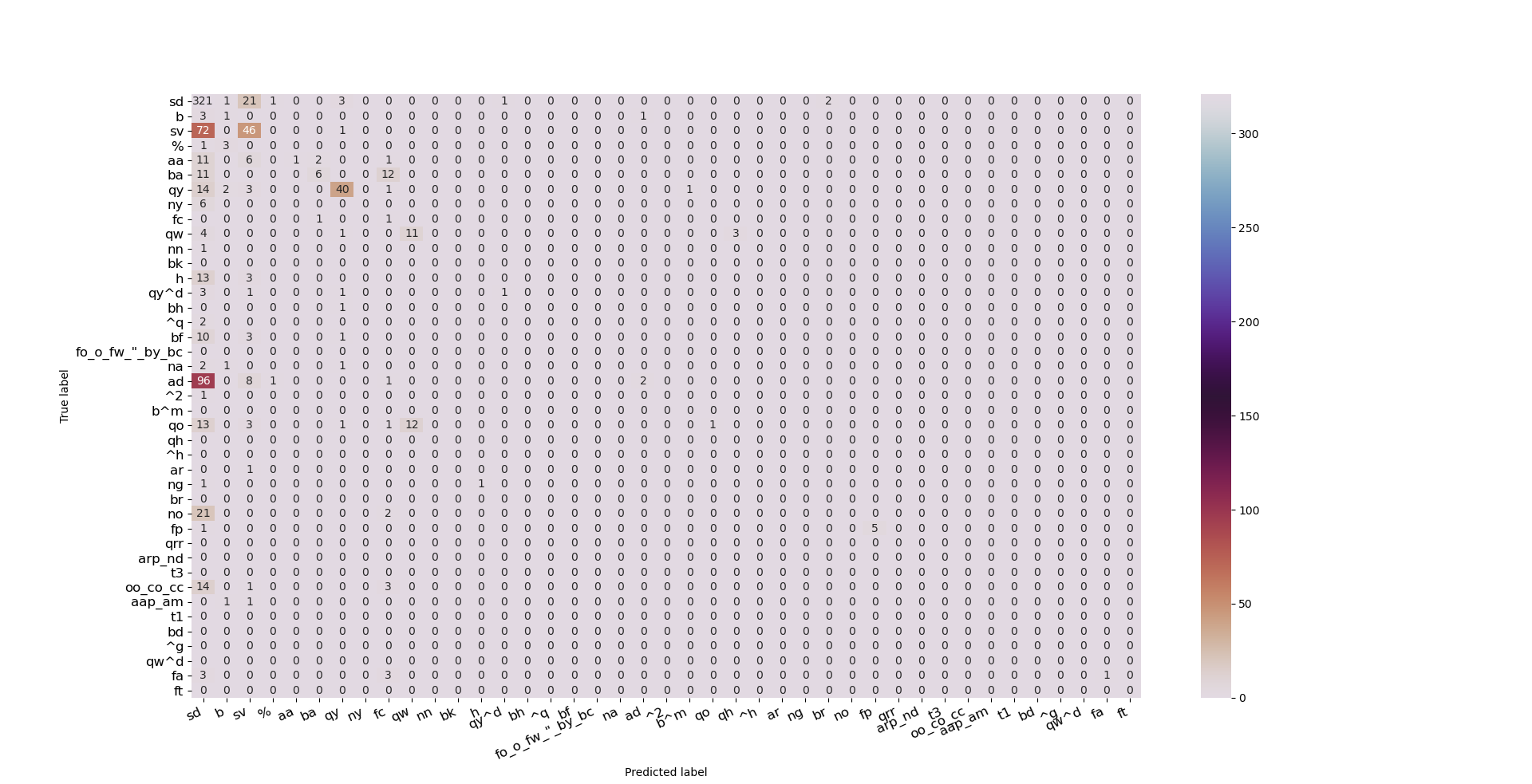}
  \caption{Confusion Matrix: Universal Sentence Encoder (all classes)}
  \label{fig:confUSE}
\end{figure*}
\begin{figure*}[hbt!]
\begin{center}
  \includegraphics[width=.60\textwidth]{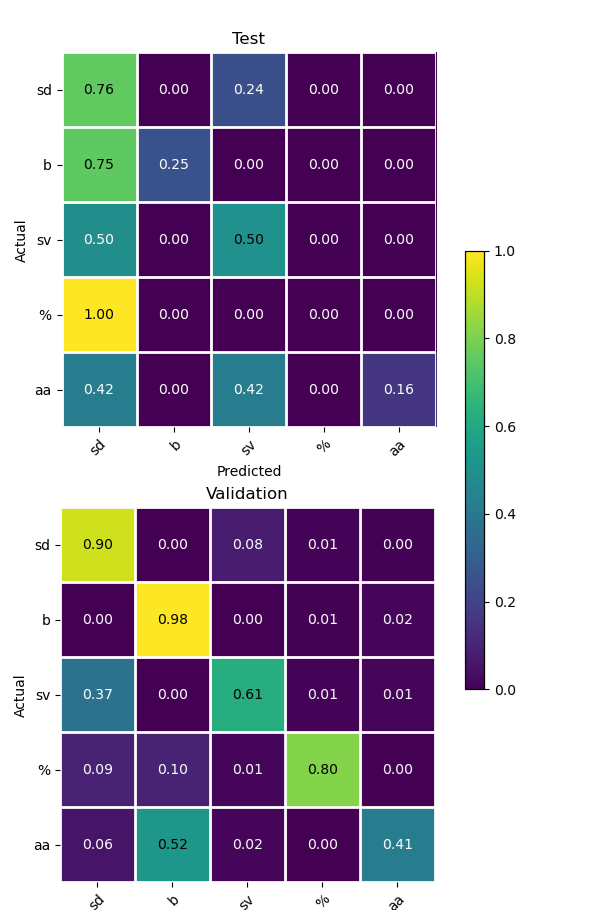}

  \caption{Confusion Matrix: Probabilistic representation+LSTM. This illustration only includes classes with the largest support. The classes shown are: sv=statement-opinion, sd=statement non-opinion, aa=agree/accept, b=acknowledge, and \%=abandoned.}
  \label{fig:confP}
 \end{center} 
\end{figure*}

It is instructive to examine the performance differences between the best (USE) and second best (probabilistic representation).
Figure \ref{fig:confUSE} shows the confusion matrix of the classification results obtained using the USE model, and Figure \ref{fig:confP} shows the confusion matrix of the probabilistic representation method. In both cases the most confused tag pair is sd (statement-non-opinion) \& sv (statement-opinion).  USE correctly classified 91.71\% of the sd occurrences, while the  probabilistic representation method only classified 76\% correctly. On the other hand USE classified 38.98\% sv utterances correctly while the probabilistic representation method classified 50\% of them correctly.  

\begin{table}
\centering
  \caption{Training, validation \& test accuracy of all the models}
  \label{tab:resultCom}
  \begin{tabular}{|c|c|c|c|}
  
    \hline
    model&acc&val\_acc&test\_acc(GitHub)\\[3ex]
    \hline
    
    GloVe+LSTM & 0.5089& 0.5195&0.3714\\[3ex]
    Prob+LSTM & 0.7672 & 0.7694& 0.4412\\[3ex]
    USE & 0.7247 & 0.6951& 0.5071\\[3ex]
    USE+LSTM & 0.3841 & 0.4257& 0.4074\\[3ex]
    BERT & 0.7151 & 0.7151& 0.4063\\[3ex]
  \hline
\end{tabular}
\end{table}
Table \ref{tab:result} shows the Precision, Recall, \& $F_1$ Score of our best performing architecture. In Table \ref{tab:result}, support represents the number of occurrences of each tag in the GitHub dataset.  USE failed to classify the third most frequent tag i.e. ad (action directive) in the test dataset. The average precision of USE over all tags is 53\%. The average recall is 51\%, and the average $F_1$ score is 42\%. A difference of only 2\% between precision and recall shows that the results of the model are consistent. 
BERT is one of the newest models for transfer learning; however our results show that fine-tuning BERT doesn't improve performance much in comparison to the Universal Sentence Encoder.  Prior work has shown that BERT does not benefits as much from fine-tuning as other embeddings~\cite{yu2019midas}. 
\begin{table}[ht]
\centering
  \caption{Precision, Recall, \& $F_1$ score of all the tags (USE)}
  \label{tab:result}
  \begin{tabular}{p{2cm}p{1cm}p{.5cm}p{1cm}p{1cm}}
  \hline
    Tag&Precision&Recall&$F_1$ score&Support\\[1ex]
    \hline
    sd& 0.51&0.92&0.66&350\\[.5ex]
     b  &     0.11  &    0.20  &    0.14 &        5\\[.5ex]
             sv &      0.47  &    0.39 &     0.43&       119\\[.5ex]
             \% &      0.00 &     0.00&      0.00   &      4\\[.5ex]
             aa &      1.00&      0.05 &     0.09  &      21\\[.5ex]
             ba &      0.67 &     0.21 &     0.32 &       29\\[.5ex]
             qy &      0.80&      0.66&      0.72 &       61\\[.5ex]
             ny&       0.00 &     0.00  &    0.00&       6\\[.5ex]
             fc&0.04&0.50&0.07&2\\[.5ex]
             qw   &    0.48 &     0.58 &     0.52  &      19\\[.5ex]
             nn &      0.00 &     0.00 &     0.00  &       1\\[.5ex]
             bk &      0.00  &    0.00&      0.00  &       0\\[.5ex]
              h&       0.00  &    0.00 &     0.00 &       16\\[.5ex]
           qy$\hat{}$d&       0.50 &     0.17 &     0.25  &       6\\[.5ex]
             bh &      0.00 &     0.00 &     0.00  &       1\\[.5ex]
             $\hat{}$q &      0.00 &     0.00 &     0.00  &       2\\[.5ex]
             bf &      0.00 &     0.00 &     0.00  &      14\\[.5ex]
fo\_o\_fw\_\"\_by\_bc &      0.00 &     0.00 &     0.00  &       0\\[.5ex]
             na&       0.00  &    0.00 &     0.00 &        4\\[.5ex]
             ad &      0.67  &    0.02 &     0.04 &      108\\[.5ex]
             $\hat{}$2&       0.00  &    0.00 &     0.00&         1\\[.5ex]
            b$\hat{}$m&       0.00 &     0.00 &     0.00 &        0\\[.5ex]
             qo &      1.00 &     0.03 &     0.06 &       31\\[.5ex]
             qh &      0.00 &     0.00 &     0.00 &        0\\[.5ex]
             $\hat{}$h &      0.00 &     0.00 &     0.00  &       0\\[.5ex]
             ar &      0.00 &     0.00 &     0.00  &       1\\[.5ex]
             ng &      0.00 &     0.00 &     0.00  &       2\\[.5ex]
             br&       0.00 &     0.00 &      0.00&         0\\[.5ex]
             no&       0.00 &     0.00 &     0.00 &       23\\[.5ex]
             fp &      1.00 &     0.83 &     0.91 &        6\\[.5ex]
            qrr &      0.00 &     0.00 &     0.00 &        0\\[.5ex]
         arp\_nd &      0.00 &     0.00 &     0.00 &        0\\[.5ex]
             t3 &      0.00 &     0.00 &     0.00 &        0\\[.5ex]
       oo\_co\_cc &      0.00 &     0.00 &     0.00 &       18\\[.5ex]
         aap\_am &      0.00 &     0.00 &     0.00 &        2\\[.5ex]
             t1 &      0.00&      0.00&      0.00&         0\\[.5ex]
             bd &      0.00 &     0.00 &     0.00 &        0\\[.5ex]
             $\hat{}$g &      0.00 &     0.00  &    0.00 &        0\\[.5ex]
           qw$\hat{}$d &      0.00 &     0.00 &     0.00 &        0\\[.5ex]
             fa&       1.00 &     0.14 &     0.25 &        7\\[.5ex]
             ft&       0.00&      0.00 &     0.00 &        0\\[.5ex]
             \hline
             &&&&\\
    avg / total &      0.53&      0.51 &     0.42  &     859\\
  \hline
\end{tabular}
\end{table}
\section{Conclusion}
This paper demonstrates a dialogue act classification system for GitHub issue comments. Due to the lack of publicly available training sets of formal teamwork dialogues, we formulated the problem as a transfer learning task, using both sentence-level and word-level embedding models to leverage information from the SwDA dataset. A significant contribution of our work is identifying the embedding model that performs best after fine-tuning on issue comments. We used GloVe, probabilistic representation, USE, and BERT embedding to train five different models. USE showed the best performance with an accuracy of 50.71\%. The low accuracy of USE on DA classification as compared to its accuracy on other state-of-the-art NLP tasks shows the complex nature of the dialogue act classification.  We evaluated many different settings for learning rates, epochs, and batch size; even though minor accuracy improvements were achievable, the performance of the embedding models remained fairly stable. 

 Our aim is to map issue comments to cognitive states in the Macrocognition in Teams Model (MITM)~\cite{fiore2010}.  Drawing from research on externalized cognition, team cognition, group communication and problem solving, and collaborative learning and adaptation, MITM provides a coherent theoretically based conceptualization for understanding complex team processes and how these emerge and change over time. MITM consists of five components: Team Problem-Solving Outcomes, Externalized Team Knowledge, Internalized Knowledge, Team Knowledge Building, and Individual Knowledge Building.
It captures the parallel and iterative processes engaged by teams as they synthesize these components in service of team cognitive processes such as problem solving, decision making and planning. MITM has been applied to other team problem solving scenarios in military logistics~\cite{hutchins2010} and business planning~\cite{seeber2013} but has never been used to analyze software engineering teams. Its usage in the domain of software engineering would be a major research contribution to the field of team cognition.
 
 Although it is possible to directly label issue comments using an MITM code book, that type of labeling would be less compatible with existing dialogue act datasets.  Instead we are constructing a mapping that relates the DAMSL tagset to these cognitive states.  For instance, the question tags in DAMSL clearly relate to information gathering processes. Also many of the DAMSL classes are less relevant to the team cognition process and could be ignored.  The most commonly occurring classes in the GitHub issue comments (statement-opinion, statement-non-opinion, agree/accept, acknowledge, and abandoned) are all relevant to the Macrocognition in Teams Model, and we plan to tune our dialogue act classifiers to bolster the performance on these classes. In future work, we continue to improve the size and quality of our publicly-released dataset by recruiting more annotators to help with the labeling task and also more systematically studying inter-coder reliability. 

\bibliographystyle{IEEEtran}

\bibliography{main}

\end{document}